\documentclass{article}
\usepackage{ijcai23}
\usepackage{pifont}
\usepackage{times}
\usepackage{soul}
\usepackage{url}
\usepackage[hidelinks]{hyperref}
\usepackage[utf8]{inputenc}
\usepackage[small]{caption}
\usepackage{graphicx}
\usepackage{amsmath}
\usepackage{amsthm}
\usepackage{amssymb}
\usepackage{booktabs}
\usepackage{algorithm}
\usepackage{algorithmic}
\usepackage[switch]{lineno}
\usepackage{enumitem}
\setlist{leftmargin=5mm}
\usepackage{threeparttable}  
\usepackage{sidecap}

\newcommand{\etal}{\textit{et al.}}
\newcommand{\xmark}{\ding{55}}
\newcommand{\cmark}{\ding{51}}%

\title{A Survey of Explainable AI in Deep Visual Modeling: Methods and Metrics}

\author{
    Naveed Akhtar
    \affiliations
    The University of Western Australia
    \emails
    naveed.akhtar@uwa.edu.au
}

\begin{document}

\maketitle

\begin{abstract}
  Deep visual models have widespread applications in high-stake domains. Hence, their black-box nature is currently attracting a large interest of the research community. We present the first  survey in Explainable AI that focuses on the methods and metrics  for interpreting deep visual models. Covering the landmark contributions along the state-of-the-art, we not only provide a taxonomic organisation of the existing techniques, but also excavate a range of evaluation metrics and collate them as measures of different properties of model explanations. Along the insightful discussion on the current trends, we also discuss  the challenges and future avenues for this research direction.  
\end{abstract}

\vspace{-5mm}
\section{Introduction}
Visual computational models have widespread applications, ranging from casual use in handheld devices to high-stake tasks in forensics, surveillance, autonomous driving and medical diagnosis etc. Contemporary visual models rely heavily on deep learning, which is a black-box technology. Hence, the opacity of deep visual models is becoming a major hurdle in enabling fairness, transparency and accountability of the AI systems that rely on these models. Given the emerging  development and deployment policies for AI around the globe, it is apparent that the future of deep visual models in practice hinges on addressing their  black-box nature.  

Historically, visual modeling has inspired numerous breakthroughs in deep learning research owing to the convenient  interpretability of visual data. The same is currently fueling a high research activity in explainable deep visual modeling. Since vision research is already a proven source of many notable advances in deep learning, systematically framing this research activity can be pivotal for the future of Explainable AI (XAI). 
Whereas the existing literature provides reviews of the broader domain of XAI, it is still void of a taxonomic treatment of the techniques devised to render deep visual models more transparent. This paper fills this critical research gap by making the following important contributions.

  \begin{figure}[t]
    \centering
    \includegraphics[width = 0.37\textwidth]{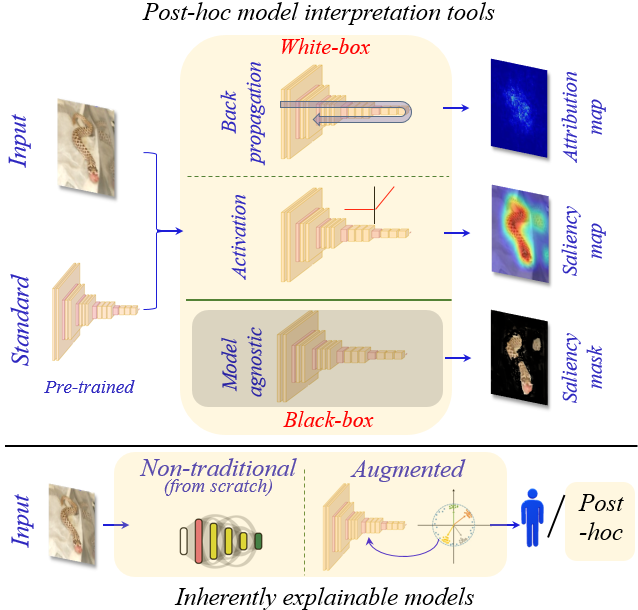}
    \vspace{-2mm}
    \caption{The literature in visual model explanation is currently dominated by post-hoc techniques (\textbf{Top}) that treat pre-trained models under black- and white-box setups. Backpropagated gradients and model activations are common tools to generate visual explanation in white-box setup, whereas model-agnostic techniques normally query the models to compute a mask for salient input regions in black-box settings. (\textbf{Bottom}) Techniques inducing  inherently explainable models rely on non-traditional building blocks for neural modeling or augment the standard blocks with interpretable modules. Their outputs may still require post-hoc tools for explanation.}
    \label{fig:teaser}
    \vspace{-2mm}
\end{figure}

\begin{itemize}[noitemsep]
    \item It provides the first comprehensive review of the literature in XAI that focuses on deep visual models\footnote{We only focus on the natural image domain, which is also the source of methods for other image domains, e.g., medical imaging.}. The review covers the influential techniques developed at the advent of the modern deep learning era, as well as the most recent state-of-the-art methods appearing in the flagship research sources of the domain. It pays a special attention to the landmark contributions. The literature is taxonomized following the technical methodologies adopted by the approaches. Thereby, providing a clear picture of the broad strategies leveraged by the existing techniques.
    \item We find that quantification of visual model interpretation performance still lacks the desired clarity in the literature due to the intrinsic difficulties in measuring the abstract subjective notion of explainablity. To help the research community addressing that, we excavate a range of evaluation metrics from the literature, and organize them as quantitative measures of different properties of explanation. This forms the second major part of our literature review. It is envisaged that this first-of-its-kind  broad methodical treatment of the evaluation metrics will significantly contribute towards the much needed systematic evaluation of the future techniques.  
    \item Due to the absence of a focused review, the current literature in XAI for visual modeling suffers from obvious inconsistencies in technical terminologies and providing the correct context to the techniques. We address this by providing accessible understanding of the terminologies along with insightful discussions throughout the paper.
    \item The paper finally highlights major open challenges for  this direction and provides the future outlook by building on the insights from the reviewed literature.  
\end{itemize}







\vspace{-3mm}
\section{Overview and Terminologies}
This section introduces the common terminologies used in the reviewed literature and formalizes the problem from the perspective of systematically categorizing the existing contributions. The existing literature is often found referring to the same concepts using different terminologies. This section also collates  these terminologies for the community.  

We find that the visual modelling domain is not too particular about discriminating between the terms \textit{explanation} and \textit{interpretation}. Without any obvious  reasons, the eventual outcomes of the methods are sometimes referred to as \textit{explanations}, whereas the same outcomes are called \textit{interpretations} in other cases. This survey favours the term \textit{explanation} for the output of a  technique. Notably, the works contributing towards the intrinsic transparency of the models generally prefer the term \textit{explainable} for the models. Methods aiming for intrinsically explainable models  pursue their goals as discussed below. 

Let $\mathcal{M}:  \mathcal{M} (\boldsymbol I) \rightarrow  \boldsymbol y$ be a visual model that maps an image $\boldsymbol I \in \mathbb R^{h \times w \times c}$ having `$c$' channels to the output $\boldsymbol y$. In   vision  literature, the explanation task mainly considers classifiers as the target models, where $\boldsymbol y \in  \mathbb R^K$ is a K-class prediction vector. Methods seeking to improve the \textit{inherent} explainability of the models see them as a hierarchy of functions, i.e., $$ \mathcal{M}(\boldsymbol I)\!\! = \!\! f_{L}(\Theta_L, f_{L-1}(\Theta_{L-1}, f_{L-2}...(\Theta_2, f_1(\Theta_1, \boldsymbol I))...) \!\! \rightarrow \!\! \boldsymbol y$$
where $\Theta_{\ell}$ denotes the model parameters (including biases) for the $\ell^{\text{th}}$ function $f_{\ell}$. These  methods  take one of two approaches. \textit{(i)}~Directly use easy to interpret functions for all  $f_{\ell}$ - see \S~\ref{sec:fromScratch}, or modify some $f_{\ell}$ for more transparency -  see \S~\ref{sec:augment} (Internal component augmentation). Models implementing this strategy are sometimes also called \textit{self-explainable} models.  \textit{(ii)}~Plug in an external \textit{explainer} module $\mathcal E(\mathcal M, \boldsymbol I)$ in the function hierarchy to generate the explanation - \S~\ref{sec:augment} (External component augmentation). This explainer is induced along the model. Collectively, the approaches under both \textit{(i)}\&\textit{(ii)} are referred to as \textit{ante-hoc} methods, as opposed to the \textit{post-hoc} techniques, which are discussed next.

The post-hoc methods consider $\mathcal{M}$ to be fixed, i.e., they do not interfere with the design or induction process of the model. To understand the common post-hoc visual explanation objective,  consider a set $\mathcal P_I = \{p_I^1, p_I^2,...,p_I^{w\times h}\} \subset \mathbb R^c$ that  contains the pixels  of the input $\boldsymbol I$. The explanation  objective here is to seek an ordered array $\mathcal W_I \subset \mathbb R$, s.t.~$|\mathcal W_I| = |\mathcal P_I|$ and the $i^{\text{th}}$ element of this array, i.e.~$w_I^i \in \mathcal W_I$, encodes a  weight for the corresponding element in $\mathcal P_I$. The weights in $\mathcal W_I$ are intended to be human-understandable. The post-hoc explanation methods generally implement the function $\mathcal S : \mathcal S (\mathcal M, \boldsymbol I) \rightarrow \mathcal W_I$. Here, $\mathcal{W}_I$ can be a binary array that represents a mask for $\boldsymbol I$ that suppresses the irrelevant $p_I^i \in \mathcal P_I$ in the transform $\mathcal{M}(\boldsymbol I) \rightarrow \boldsymbol y$. This objective is more pertinent to the \textit{model-agnostic} post-hoc methods - see \S~\ref{sec:ModelAgnostic}. These methods avoid using any internal signal of the model, e.g., neuron activations, thereby assuming a \textit{black-box} setup which does not allow  access to the model details.  

When access to the model is available, the weights $w_I^i \in \mathcal W_I$ take real values to quantify the importance of the input pixels. To that end, a wide range of methods rely mainly on the activations of the model's internal neurons to estimate $\mathcal W_I$. We call such methods \textit{neuron activation-based} techniques - see \S~\ref{sec:sali}. The output of such methods is  commonly known as a \textit{saliency map}. The activation-based techniques often also use model gradients. However, those gradients are generally not backpropagated right until the input. The methods that compute the model gradients w.r.t.~the input are categorised as the \textit{backpropagation-based} methods in this survey - see \S~\ref{sec:BPm}. The output of these methods is more commonly termed \textit{attribution map}. Note that, attribution maps and saliency maps are technically `synonyms'. We find multiple papers in this domain that use these terms interchangeably to describe the generated visual explanations. Other common terms that refer to the same concept in the literature are \textit{relevance} scores,  feature \textit{contributions} and  \textit{importance} maps.  

Whereas the ante-hoc methods render the model intrinsically more transparent, the  interpretation process may still use post-hoc tools or need human domain experts to explain the  predictions. Within the vision domain, the preferred mode of explanations are visualizations. However, rarely, we also encounter text as the explanations~\cite{hendricks2016generating}, \cite{park2018multimodal}. We also find that the common objective of the post-hoc techniques is to generate \textit{input-specific} explanations. A rare exception to that is \textit{input-agnostic saliency mapping} \cite{akhtarAAAIrethinking} that generates visual explanations that encode generic understanding of the model of human-defined concepts. 



\vspace{-3mm}
\section{Inherently explainable models}
\vspace{-1mm}
We find that there is a prevailing impression in the community that explainability and performance of deep visual models are inversely correlated. However, Rudin~\shortcite{rudin2019stop} rightly noted a lack of concrete evidence in this regard, and argued in favor of developing inherently explainable models. To that end, recent papers devise  approaches  along two broad directions: \textit{(a)}~\textit{designing explainable models from scratch}, and \textit{(b)} \textit{augmenting black-box models with interpretable components}.  

\begin{SCfigure*}
    \centering
    \includegraphics[width = 0.77\textwidth]{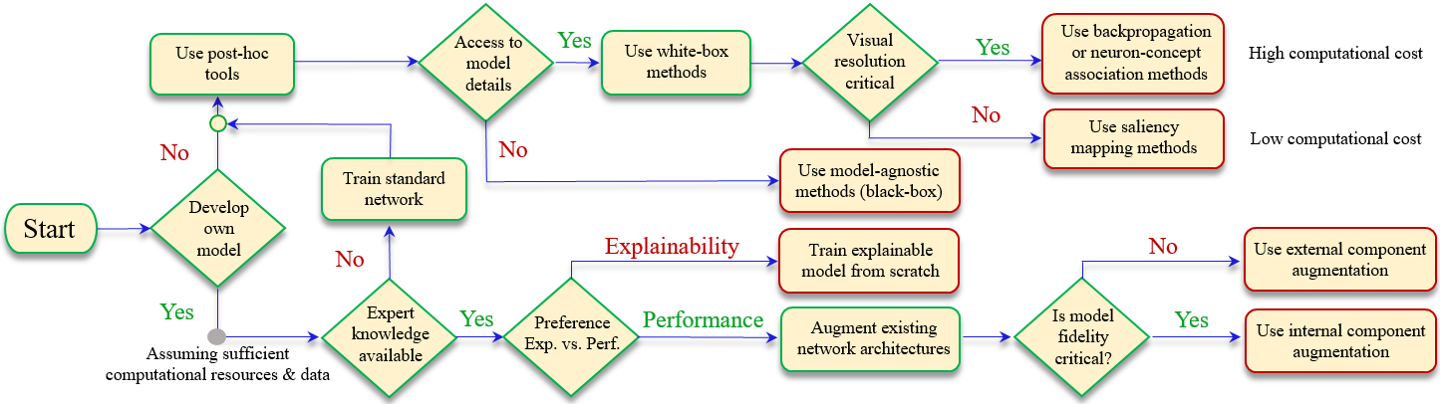}
    \vspace{-1mm}
    \caption{Flowchart for general guidance to select explanation tools for visual model application domains. The chart follows terminologies used to categorise the literature in this survey. Best viewed enlarged.}
    \label{fig:flow}
    \vspace{-3mm}
\end{SCfigure*}

\vspace{-2mm}
\subsection{Explainable model design from scratch}
\label{sec:fromScratch}
\vspace{-0.75mm}
Along this direction, a category of methods explores the lore of generalized additive models (GAMs)~\cite{hastie2017generalized}. The central idea is to implement the intended transformation (i.e., modeling)  of the input as a set of additive sub-models that deal with the input features separately. This is used to  preserve the interpretability of the overall transformation~\cite{agarwal2021neural}, \cite{chang2021node}, \cite{dubey2022scalable}, \cite{alvarez2018towards}.
However, the lack of expressive power of such composite modelling, and its limited scalability are major hurdles in the practicability of this strategy. Taking a different approach, Blazek  
 \etal~\shortcite{blazek2021explainable} proposed  essence neural networks that employ non-standard building blocks of  Support Vector Machines based neurons. With a systematic architectural design of the underlying neural network, they showed remarkable generalizable abilities of the approach. However, the design process of their network remains cumbersome and relies heavily on an active involvement of the domain experts. 
 Li \etal~\shortcite{li2021scouter} proposed to build the classification stage of a visual classifier using a slot attention mechanism that learns a semantically meaningful internal representation. However, their method must still rely on a backbone module in the early stage of the network, which compromises the explainability of the overall model. 

It is a common knowledge that a key strength of deep learning, especially in the vision domain, is its ability to represent the data with discernible low- and high-level features through a hierarchical structure. Methods aiming at fully explainable models must account for such a sensible hierarchical treatment of the domain data in the network structure manually. This generally requires  expert knowledge of the domain, which makes fully explainable modeling a less attractive option for many problems.        
\vspace{-2mm}
\subsection{Augmenting neural models for explainability}
\label{sec:augment}
\vspace{-0.5mm}
In general, explainability of visual models is seen as a continuum rather than a discrete measure. This perspective has encouraged researchers to devise methods for improving the `extent' of interpretability of the models. Generally, this is pursued by augmenting the conventional neural models  with \textit{(i) interpretable internal components}, or \textit{(ii) external components providing the desired  explanations}.

\vspace{-0.5mm}
\paragraph{Internal component augmentation:}
To improve the intrinsic explainability of visual models, 
Chen \etal~\shortcite{chen2020concept} introduced a \textit{concept whitening} module that aims at aligning the axes of the latent space of the model with  apriori  known concepts. This alignment also involves a process of de-correlating the concepts space for  intelligibility. Wang \etal~\shortcite{wang2021interpretable} also explored a closely related mechanism with a plug-in embedding space for the model that is spanned
by basis concepts. Here, the basis concepts are kept category-aware, and the within-category concepts
are kept orthogonal to each other for their human-understandability. These are indeed useful methods, however they come at the cost of further complicating the training process of neural modeling. 

Concept bottleneck modelling~\cite{koh2020concept} is another interesting framework that lets a neural model first learn human-interpretable (sub-)concepts, and makes the final prediction by further processing those concepts. Architecturally, it introduces a concept layer in the network that also enables human intervention during the prediction stage in \cite{chauhan2022interactive}. Concept bottleneck models (CBMs) are currently attracting notable attention of the research community. Bahadori \etal~\shortcite{bahadori2020debiasing} performed causal reasoning to debias such models, whereas Antognini \etal~\shortcite{antognini2021rationalization} used this framework for the application of textual rationalization. Working on the principles similar to CBMs, ProtoPNetwork~\cite{chen2019looks} also introduces a prototype layer in the network  that associates high-level sub-concepts to its neurons. 

Notice that some approaches discussed in \S~\ref{sec:fromScratch}  have a close conceptual resemblance to CMBs. For instance, like CBMs,  only the later layers of the visual models in \cite{blazek2021explainable} are interpretable. This is because, their technique must rely on deep features (instead of image pixels) to construct the explainable (sub-)network in the case of visual modeling. However, unlike \cite{blazek2021explainable}, the techniques using internal component augmentation employ  the standard neurons along some basic arithmetic operations to improve the  interpretability of the network layer(s). It is notable though,  both paradigms remain interested in the \textit{deeper} layers of the neural network for model explainability. 
This is true even for \cite{kim2022vit}, which specifically attempts to make Vision Transformers 
more explainable by using neural trees as a decoder of the representation learned by the model, where the nodes of the trees are contextual Transformer modules. 

\vspace{-0.5mm}
\paragraph{External component augmentation:}
Internal component augmentation techniques embed modules in the neural network such that they do not form a new branch of the graph. In contrast, some recent techniques  introduce explanation modules as external branches of the network. For instance, 
Sarkar \etal~\shortcite{sarkar2022framework} proposed a framework to learn an external explanation module in the context of concept-based models~\cite{koh2020concept}. Their explanation module uses an external concept decoder to improve model interpretability.  Similarly, \cite{lang2021explaining} trains an external StyleGAN as an explainer for a visual classifier. Human-understandable explanations are produced in this approach by relying on semantically meaningful dimensions learned by the style-space of the external module. Another example of external explainer can be found in \cite{stalder2022you}, which is claimed to produce sharp masks of relevant image regions using a trainable model. 
Though useful, an apparent drawback of the external module-based methods is that the module can inadvertently affect both the model performance and its explanations. For the later case, the explanation can unintentionally be encoding the module's  behavior instead of explaining the model.  
    

\vspace{-2mm}
\section{Post-hoc explanations}
\label{sec:postHoc}
\vspace{-0.5mm}
It is hard to  identify any study  that rigorously establishes an inverse correlation between the model explainability and its performance. Nevertheless, it is widely presumed that this inverse relation exists. Consequently, the research community is relatively more active in devising post-hoc explanation tools as compared to developing intrinsically explainable models.  
Post-hoc tools include a category of methods that leverage the model information when it is available - see \S~\ref{sec:sali} and \ref{sec:BPm}, whereas another class of methods can handle the scenario when such information is unavailable - see \S \ref{sec:ModelAgnostic}. This flexibility, along the fact that these methods do not interfere with model performance, makes post-hoc tools very practical. 



\vspace{-1.5mm}
\subsection{Model-agnostic methods}
\label{sec:ModelAgnostic}
For the cases where the model information is unavailable, i.e., black-box scenarios, post-hoc explanation methods query the model in some form to identify the important pixels in the input in a model-agnostic manner. For instance, Fong \etal~\shortcite{fong2019understanding} used queries to compute extremal perturbations to  identify the most salient input pixels.  
 Local Interpretable Model-agnostic Explanations (LIME) is an early attempt of model-agnostic methods that approximates the  model behavior within a local neighborhood around the input by fitting a simple interpretable model to the perturbed versions of the input~\cite{ribeiro2016should}. However, LIME is known to be unstable. Dhurandhar \etal~\shortcite{dhurandhar2022right} recently proposed a fix to this instability. Randomized Input Sampling for Explanation (RISE)~\cite{petsiuk2018rise} is yet another popular model-agnostic explanation method. Lundberg \& Lee~\shortcite{lundberg2017unified} proposed SHapley Additive
exPlanations (SHAP) that leverage Shapely values~\cite{shapley1997value} to estimate the contribution of each feature to the prediction. Recently, it is claimed that the marginal contribution computed by the Shapely values is sub-optimal for model explanation. Hence, a weightedSHAP is introduced in \cite{kwon2022weightedshap}.   

Model-agnostic strategy is the only scheme available for explaining a model in a black-box setup. 
Since the approaches do not leverage model information, they solve a harder problem to generate explanations. Thus, we often find these methods relying on heuristics to control their solution space. These heuristics can have unintended effects on the generated explanations. In general, despite using heuristics, these methods remain  computationally expensive. 

When the internal information of the model is available, there are two  important  signals that can be exploited to explain the model behavior; namely, \textit{(i) internal activations} and (\textit{ii) backpropagated gradients}. We discuss the methods exploiting \textit{(i)} and \textit{(ii)} in \S~\ref{sec:sali} and \S \ref{sec:BPm}, respectively. 

\vspace{-1.5mm}
\subsection{Neuron activation-based techniques}
\label{sec:sali}
\vspace{-0.75mm}
 There is a range of methods that strongly rely on the activation values of the internal neurons of the model to explain their  predictions. It is emphasized though that the activations may not be the only internal signals used by these methods. 

 \vspace{-0.75mm}
\paragraph{Saliency methods:}
A popular stream of techniques computes \textit{saliency maps} for the inputs as visual explanations using  internal neuron activations of the models. 
Class Activation Mapping (CAM)~\cite{zhou2016learning} is one of the earliest methods in this direction, which employs the global average pooling neuron activations to weakly localize the objects in the inputs for saliency mapping. Oquob \etal~\shortcite{oquab2015object} used the max pooling neurons for a similar purpose. Whereas multiple other techniques exist for activation-based saliency mapping, GradCAM~\cite{selvaraju2017grad} is a landmark work, which  combines the model gradients with the activations of its internal neurons to compute the saliency maps. Multiple variants of GradCAM have appeared in the literature, e.g., GradCAM++ \cite{chattopadhay2018grad}, Score-CAM~\cite{wang2020score}, Smooth-GradCAM++ \cite{omeiza2019smooth}, all of which use neuron activation signals. Generally, there is a large disparity between the sizes of the activation maps of the internal layers of  a model and the input. This requires interpolation of the activation signals for computing the eventual saliency map, which results in poor resolution of the maps. Recently, Jalwana \etal~\shortcite{jalwana2021cameras} addressed this weakness with their CAMERAS method that employs a multi-scale mapping for high-resolution saliency mapping. 

Despite a considerable number of existing methods for  activation-based saliency mapping, we still regularly witness new techniques emerging along this line of research. 
On the other hand, there are still many other popular classic techniques, e.g., Layerwise Relevance Propagation (LRP)~\cite{bach2015pixel}, DeepLift \cite{shrikumar2017learning} that rely on activations of multiple layers to compute the saliency maps. With the paradigm shift in visual modeling in favor of Transformers, these classical methods are also being extended to explain the Transformer based models~\cite{ali2022xai}.  

\vspace{-0.75mm}
\paragraph{Neuron-concept association methods:}
Whereas techniques like CBMs \cite{koh2020concept} encourage neuron-concept association by design,  we also find methods that aim at identifying such an association in a post-hoc manner. For instance, Bau \etal~\shortcite{bau2017network} aimed at quantifying the alignment between the model neurons and known semantic concepts.  Similarly, Mu \etal~\shortcite{mu2020compositional}  searched for logical  forms of explanation defined by a composition of the concepts learned by polysemantic neurons.  A neuron-concept explainer is also developed in \cite{wang2022hint} that utilizes both saliency maps and Shapley values. Previously, Shapley values of neurons are also leveraged by  \cite{ghorbani2020neuron} to analyze the importance of neurons for a given output. On the other hand,  activations are also used by the popular TCAV method~\cite{kim2018interpretability} to identify the sensitivity of the neurons to a given set of concept. 

\vspace{-1.5mm}
\subsection{Backpropagation-based methods}
\label{sec:BPm}
Gradients of a model w.r.t.~input are 
widely considered a natural analogue of its coefficients in the modern deep neural networks. We find a range of methods leveraging backpropagated model gradients for explanation.  Simonyan \etal~\shortcite{simonyan2013deep} were among the first to estimate input feature relevance for the prediction with model gradients.  Guided-backpropagation \cite{springenberg2014striving}, smoothGrad \cite{smilkov2017smoothgrad}, fullGrad~\cite{srinivas2019full} and multiple other methods in the literature later that used the backpropagated gradients to estimate the feature attribution scores. It is noteworthy that  backpropagated gradients are sometimes also employed by other categories of methods. For instance, the saliency methods, e.g., Grad-CAM, CAMERAS, in \S~\ref{sec:sali} also use model gradients. However, those gradients are not fully propagated back to the input. 

Within the methods mainly relying on the backpropagated gradients, there is an important sub-category of \textit{path attribution methods}. Inspired by game-theory, these methods accumulate the backpropagated gradients of the images defined by a path between the input  and a reference image, where the latter signifies the absence of the features in the input. The pioneering work by Sundrarajan \etal~\shortcite{sundararajan2017axiomatic} claims  that the path attribution strategy can preserve certain desirable axiomatic properties for the explanations. Their claims were further qualified recently in \cite{lundstrom2022rigorous}. The original work has inspired a range of methods employing the path attribution framework~\cite{yang2023local}, \cite{erion2021improving}. However, Stumfels \etal~\shortcite{sturmfels2020visualizing} have also noted that the path attribution strategy can suffer from certain artifacts resulting from the reference image itself. Our close inspection revealed that despite their axiomatic nature, these methods sometimes compute counter-intuitive results. Moreover, the literature often fails to establish that the eventual methods actually retain the claimed properties.         



\begin{table*}
  \centering
 \caption{Representative evaluation metrics and their ability to measure the extent of different properties of explanations. More *s indicate better ability. Refer to \S~\ref{sec:EM} for the description of the properties. GT denotes Ground Truth usage, and Comput.~is computational efficiency.}
 \vspace{-3mm}
  \resizebox{1\textwidth}{!}{
    \begin{tabular}{|l|c|c|c|c|c|c|c|c|c|}
    \hline
    \multicolumn{1}{|c|}{\textbf{Metric}} & \textbf{Work} & \textbf{Model Fidelity} & \textbf{Localisation} & \textbf{Stability} & \textbf{Conciseness} & \textbf{Sanity preservation} & \textbf{Axiomatic properties} & \textbf{GT} & \textbf{Comput.} \\
    \hline \hline
    Pixel Flipping & Bach \etal~\shortcite{bach2015pixel} & ****  & *     & \xmark  & *     & **    & \xmark  & \xmark  & *** \\
    \hline
        SensitivityN & Ancona \etal~\shortcite{ancona2017towards} & ****  & *     & \xmark  & *     & **    & ***   & \xmark  & * \\
    \hline
    Insertion/deletion & Petsiuk \etal~\shortcite{petsiuk2018rise} & ****  & *     & \xmark  & *     & **    & \xmark  & \xmark  & *** \\
    \hline
    ROAR  & Hooker \etal~\shortcite{hooker2019benchmark} & ****  & *     & \xmark  & *     & **    & \xmark  & \xmark  & * \\
    \hline
    ROAD  & Rong \etal~\shortcite{rong2022consistent} & ****  & *     & \xmark  & *     & **    & \xmark  & \xmark  & ** \\
    \hline
    Pointing game & Zhang \etal~\shortcite{zhang2018top} & \xmark  & ****  & \xmark  & *     & \xmark  & \xmark  & \cmark & *** \\
    \hline
    Stability  & Alvarez \etal~\shortcite{alvarez2018towards} & **    & \xmark  & ****  & \xmark  & *     & \xmark  & \xmark  & ** \\
    \hline
    Sensitivity & Yeh \etal~\shortcite{yeh2019fidelity} & **    & \xmark  & ****  & \xmark  & *     & \xmark  & \xmark  & ** \\
    \hline
    Sparseness & Chalasani \etal~\shortcite{chalasani2020concise} & *     & *     & \xmark  & ****  & *     & \xmark  & \xmark  & ** \\
    \hline
    Effective complexity & Nguyen \etal~\shortcite{nguyen2020quantitative} & *     & *     & \xmark  & ****  & *     & \xmark  & \xmark  & ** \\
    \hline
    Cascading randomization & Adeboya \etal~\shortcite{adebayo2018sanity} & ***   & *     & \xmark  & \xmark  & ****  & \xmark  & \xmark  & ** \\
    \hline
    Completeness & Sundararajan \etal~\shortcite{sundararajan2017axiomatic} & **    & \xmark  & \xmark  & *     & **    & ****  & \xmark  & * \\
    \bottomrule
    \end{tabular}%
    \label{tab:addlabel}
    }
    \vspace{-4mm}
\end{table*}%


\vspace{-2mm}
\section{Evaluation metrics}
\label{sec:EM}
\vspace{-0.75mm}
Evaluating the performance of a model explanation tool is not straightforward. Due to the abstract subjective nature of the notion of explanation and the lack of ground-truth, objectively evaluating a method's performance  requires quantifying multiple aspects of the computed explanation. 
Hence, development of metrics to evaluate explanation methods is currently actively pursued. We excavate numerous metrics from different contributions and organise them based on the explanation properties quantified by them.     

\vspace{-1.5mm}
\subsection{Quantifying the model fidelity}
\label{sec:ModFid}
Fidelity of the explanation to the model is a measure of the extent to which the explanation captures the true behavior of the model. It is the most essential property of the computed explanation. Broadly, the metrics quantifying this property  rely on estimating some form of correlation between the relevance of the input features (as computed by the explanation method) and the output variation when those features are perturbed/removed. The core intuition is that for a model-faithful explanation, the output variations should be stronger for the perturbations to the more relevant features.

As an early example of evaluating the model fidelity, Bach \etal~\shortcite{bach2015pixel} developed a  metric of \textit{pixel flipping} where the fidelity is evaluated by observing the change in the prediction scores by flipping the input pixels with high (absolute) scores. The metric was later generalized to \textit{region perturbation} in \cite{samek2016evaluating} where a random sample from a uniform distribution is used to perturb a set of input pixels to analyze the effects on predictions. Similarly,  Alvarez \etal~\shortcite{alvarez2018towards} removed the pixels from the input and measured the correlation between the prediction scores and 
 the attribution scores. 
In \cite{ancona2017towards}, \textit{SensitivityN} metric is proposed
to measure the correlation between the sum of a subset of computed pixel attributions and the change in the prediction when the corresponding pixels are perturbed. This notion is further generalized in \cite{yeh2019fidelity} with an \textit{Infidelity} measure as the Expected difference between the dot product
of the input perturbation to the attribution, and the prediction change resulting from the perturbation.

The key strategy of the above metrics is to perturb (or remove) input features and analyze the effects on the output. 
However, Hooker \etal~\shortcite{hooker2019benchmark} showed that this results in a distribution shift problem for the models because their training process does not account for any pixel removal or perturbation. This compromises the reliability of the evaluation metrics. Hence, they proposed to iteratively retrain the model after the removal of each pixel on the modified training data. Their Remove and Retrain (\textit{ROAR}) is one of the widely adopted  metric due to its global (in terms of training data) nature. Nevertheless, its evaluation can be computationally prohibitive. To address that, Rong \etal~\shortcite{rong2022consistent} proposed a Remove and Debias (\textit{ROAD}) metric that avoids the retraining with a de-biasing step. The authors claim a desirable property of \textit{consistency} between the ROAD scores when the most relevant pixels are removed first, or the least relevant pixels are removed first - a property  not exhibited by ROAR.

Whereas multiple metrics exist for model fidelity, there is still no single agreed-upon metric that is considered to accurately and comprehensively quantify this property. The emerging relevant trends include combining the existing metrics~\cite{yang2023local} and quantifying the model fidelity with other properties such as \textit{consistency} and \textit{sufficiency} of the computed attribution~\cite{dasgupta2022framework} or proposing novel settings to evaluate this property~\cite{rao2022towards}. A comprehensive and reliable evaluation of this property is still an open challenge for the vision domain. 

\vspace{-1.5mm}
\subsection{Quantifying the localisation ability}
\vspace{-0.5mm}
An accurate explanation must assign high relevance to the foreground object features. This intuition has led to a range of metrics that aim at leveraging ground truth in quantifying the explanation performance. Their key idea is to measure how well the computed relevance scores correlate with the bounding boxes, segmentation masks, or the grid cells that contain the object of interest. In that sense, we can collectively see them as localisation quantifying metrics. 

\textit{Pointing game}~\cite{zhang2018top} is one of the popular examples in this direction, which computes  the percentage of the most relevant pixel for every sample  being located on the foreground object. Similarly, Kohlbrenner \etal~\shortcite{kohlbrenner2020towards} proposed an \textit{attribution localization} metric that computes the fraction of the cumulative relevance score attributed to the foreground object in an input. The same measure is termed \textit{relevance mass} in \cite{arras2022clevr}, where another metric of  \textit{relevance rank} is also introduced, which estimates the fraction of the object mask pixels common to the top $K$ relevant pixels, where $K$ is the mask size. In \cite{arras2022clevr}, mosaics of different data samples are use to quantify the fraction of the  relevance assigned to the foreground object. 

Whereas the localisation based metrics can leverage the ground truth, their objective is agnostic to model fidelity. An explanation method enforcing stronger attribution to the foreground, irrespective of the actual behavior of the model, will achieve higher score for these metrics. Hence, relying solely on the localisation based metrics can be misleading.  

\vspace{-1.5mm}
\subsection{Quantifying the stability}
A correct explanation should not change drastically when the input is changed only slightly. Following this intuition multiple metrics have emerged that quantify some form of stability of the computed explanations. For instance, Avarez \etal~\shortcite{alvarez2018towards} measured the local Lipschitz continuity of the explanation to observe its stability. Similarly, Motavon \etal~\shortcite{montavon2018methods} computed the maximum $\ell_1$-distance between the explanations of two inputs normalized by the Euclidean distance between them, to observe the explanation stability. In \cite{yeh2019fidelity}, an estimate of the maximum distance between the explanations of the neighborhood inputs is considered to quantify the stability. 
Dasgupta \etal~\shortcite{dasgupta2022framework} quantified the probability of the similarity between the explanations of  similar samples. In \cite{agarwal2022rethinking}, relative changes in the explanations are estimated with respect to the input, model embeddings and the model output logits to propose three measures of stability, namely; \textit{relative input stability}, \textit{relative representation stability} and \textit{relative output stability} of the explanations. The authors argue to discard the assumption that the model is a black-box (for black-box methods) and propose to leverage full model information to ensure a more reliable evaluation of the explanation method.

Whereas the notion of stability is understood in the same sense across different papers, we do not find consensus in its naming convention. For instance, stability is referred to as \textit{continuity} in \cite{montavon2018methods}, \textit{sensitivity} in \cite{yeh2019fidelity} and \textit{consistency} in \cite{dasgupta2022framework}. It is also worth highlighting that the prevailing existing notion  of stability is underpinned by the assumption that the model in question is smooth. Stability measures for the explanation of a model not following this assumption can be problematic. 

\vspace{-1mm}
\subsection{Evaluating other desirable properties}
\label{sec:Otherprop}
Besides the above, we also find a number of  other metrics in the literature that address other desirable properties of the explanations. We summarise them below.  

\vspace{-0.75mm}
\paragraph{Conciseness:}
An effective explanation  should be concise. Following this intuition, Chalasani \etal~\shortcite{chalasani2020concise} quantified the explanation conciseness  by estimating its \textit{sparseness} with the Gini Index, whereas Bhatt \etal~\shortcite{bhatt2020evaluating} evaluated the 
 explanation \textit{complexity}  to measure the conciseness using a fractional contribution distribution of the input features for the explanation. Similarly, Nguyen \etal~\shortcite{nguyen2020quantitative}  also measured an \textit{effective complexity} of the explanation, where low complexity implies less reliance on more input features. Again, whereas different naming conventions are used by the original works, they are all concerned with  \textit{conciseness}. 

\vspace{-0.75mm}
\paragraph{Sanity preservation:}
Adeboya \etal~\shortcite{adebayo2018sanity} first showed that multiple saliency mapping methods fail basic sanity checks. Thereafter, sanity preservation is often used as an evaluation criterion for the saliency methods \cite{jalwana2021cameras}. \textit{Cascading randomization}~\cite{adebayo2018sanity} is the common test in this regard that alters model weights in a cascading manner to visually observe the effects of this randomization on the computed explanation. Sixt \etal~\shortcite{sixt2020explanations} also proposed a test for the backpropagation based methods which observes the effects of randomizing the later layers on the explanation. 
Such sanity preservation tests can often be viewed as convenient qualitative (in some cases, quantitative) measures of other properties, e.g., model fidelity.      

\vspace{-0.75mm}
\paragraph{Axiomatic properties:}
In the above, we are able to organize the evaluation metrics systematically. This owes to the fact that these metrics are essentially developed to evaluating specific properties of the explanation. Nevertheless, those properties themselves remain  abstract notions. In contrast, there are a few  properties that have concrete mathematical definitions, e.g., \textit{completeness}, \textit{linearity}, \textit{input invariance}~\cite{sundararajan2017axiomatic}, \cite{kindermans2019reliability}. Measuring the extent to which these axiomatic properties are satisfied by the explanations, is also employed as an evaluation criterion. However, it is more relevant to the game-theory  inspired path attribution methods~\cite{sundararajan2017axiomatic}. We refer to \cite{lundstrom2022rigorous} for the details on the  axiomatic properties. More recently, Khakzar \etal~\shortcite{khakzar2022explanations} also proposed an empirical framework that evaluates explanations using multiple non-game theoretic axioms.  

\vspace{-2mm}
\section{Challenges and future directions}
\vspace{-0.5mm}
The notion of explanation is abstract and subjective. Its ill-defined nature poses multiple challenges for the research community, especially in the vision domain where the qualitative aspects of perceptual data   confound even the objective of explanation. With this first literature review of XAI for visual models, we are able to identify the key  challenges for this direction. They are listed below with pointers to the potential avenues to explore for their solution.  


\vspace{-0.75mm}
\paragraph{Post-hoc methods not addressing model  opacity:}
The very aim of XAI for deep learning is to  address the black-box nature of the models. However, despite explaining a model prediction with the post-hoc techniques, the model retains  nearly the same level of opacity because the explanation relates only to a single sample. Current literature is dominated by  post-hoc techniques, which places a question mark on the current objective of computing \textit{input-specific} explanations pursued by these  methods. Ideally, this dominant stream of methods should also help in reducing the model opacity by explaining their \textit{general} behavior. Very recently, interesting results in this regard are reported in \cite{akhtarAAAIrethinking}, where the saliency maps are computed to capture the model's sensitivity to generic geometric patterns in an input-agnostic manner. We can expect future post-hoc techniques to explore this concept further.  

\vspace{-0.75mm}
\paragraph{Correlation vs causation:}
It is often the case that ``correlation is not causation''. The current  literature in  vision domain predominantly sees the explanation problem as computing correlation between the input features and the model prediction. For instance, the  saliency maps - see \S~\ref{sec:sali} -   simply highlight the image regions having high relevance to the model prediction. Better correlation is also the criterion to improve the model fidelity scores - see \S~\ref{sec:ModFid}. Though useful, correlation alone is insufficient to truly explain the model. On one hand, it is imperative to develop techniques that account for causation; on the other, we need evaluation metrics to measure the causal capacity of the existing methods. We anticipate these methods to be more sophisticated and evaluation metrics to be more comprehensive than the existing ones.

\vspace{-0.75mm}
\paragraph{Inspirational domain mismatch:}
Visual data has its peculiar nature. Incidentally, many  techniques for  explaining the visual models are inspired by other domains. For instance, path-based methods (in \S~\ref{sec:BPm}) and axiomatic properties evaluation (in \S~\ref{sec:Otherprop}) have game-theoretic roots. Similarly, Shapely values~\cite{shapley1997value} are Economy related concepts. The inherent mismatch between the vision domain and the inspiring  domains can lead to  problems. However, this aspect is often overlooked in the literature. For example, it is very hard to define `absence' of a feature in an image to correctly identify the `reference' for that image in    path based methods - see \S~\ref{sec:BPm}. This is not an issue in other game-theory  applications. Interestingly, we also witnessed that the eventual attribution maps computed by the existing path based methods seldom exhibit the acclaimed axiomatic properties in a strict sense. The future explorations must address such domain mismatch problems diligently to make the explanations reliable. 

\vspace{-0.75mm}
\paragraph{Performance vs transparency dilemma:} Indeed, we find a trend that more human-understandable models have lower performance. Models not accounting for explainability are currently the top performers in the  vision domain. This observation is likely a major contributor to  the common belief that explainability and performance have an inverse relation. However, we do not find rigorous studies in the current literature that analytically (or more systematically) strengthen or refute this idea. A few leading vision research labs even believe explainability to be an unnecessary trait in practice. However, regulatory authorities concerning the development and deployment of AI have a consensus that the systems need to be transparent/explainable. Resolving this conflicting situation demands a comprehensive investigation of the `performance vs transparency' dilemma. We can anticipate some research activity addressing this  in the future.  

\vspace{-0.75mm}
\paragraph{The (in)consistency of methods and metrics:}
Due to the absence of concrete ground truth for the explainability task in general, the methods are hard to evaluate. The issue is further riled by the inconsistency of the available evaluation metrics. For instance, statistical unreliability and inconsistency of a few such metrics is already proven in \cite{tomsett2020sanity}. Implying, comparative rankings of the saliency methods is untrustworthy under these metrics. At the same time, explanation methods themselves are known to sometimes fail basic sanity checks~\cite{adebayo2018sanity}. This dicephalous nature of the problem demands further research that fully focuses on each sub-problem individually. Instead of making minor contributions of evaluation metrics along proposing new explanation methods (a common trend in the literature), thorough works that fully focus on devising the evaluation metrics and schemes are needed. There is also an obvious need to invent techniques to provide ground truth for  evaluations. 

\vspace{-2mm}
 \bibliographystyle{named}
 \bibliography{ijcai23}

\begin{thebibliography}{}

\bibitem[\protect\citeauthoryear{Adebayo \bgroup \em et al.\egroup
  }{2018}]{adebayo2018sanity}
Julius Adebayo, Justin Gilmer, Michael Muelly, Ian Goodfellow, Moritz Hardt,
  and Been Kim.
\newblock Sanity checks for saliency maps.
\newblock {\em NeurIPS}, 2018.

\bibitem[\protect\citeauthoryear{Agarwal \bgroup \em et al.\egroup
  }{2021}]{agarwal2021neural}
Rishabh Agarwal, Levi Melnick, Nicholas Frosst, Ben Lengerich, Rich Caruana,
  and Geoffrey Hinton.
\newblock Neural additive models: Interpretable machine learning with neural
  nets.
\newblock {\em NeurIPS}, 2021.

\bibitem[\protect\citeauthoryear{Agarwal \bgroup \em et al.\egroup
  }{2022}]{agarwal2022rethinking}
Chirag Agarwal, Nari Johnson, Martin Pawelczyk, Eshika Saxena, Marinka Zitnik,
  and Himabindu Lakkaraju.
\newblock Rethinking stability for attribution-based explanations.
\newblock {\em arXiv:2203.06877}, 2022.

\bibitem[\protect\citeauthoryear{Akhtar and
  Jalwana}{2023}]{akhtarAAAIrethinking}
Naveed Akhtar and Mohammad Jalwana.
\newblock Rethinking interpretation: Input-agnostic saliency mapping of deep
  visual classifiers.
\newblock In {\em AAAI}, 2023.

\bibitem[\protect\citeauthoryear{Ali \bgroup \em et al.\egroup
  }{2022}]{ali2022xai}
Ameen Ali, Thomas Schnake, Oliver Eberle, Gr{\'e}goire Montavon, Klaus-Robert
  M{\"u}ller, and Lior Wolf.
\newblock Xai for transformers: better explanations through conservative
  propagation.
\newblock {\em ICML}, 2022.

\bibitem[\protect\citeauthoryear{Alvarez~Melis and
  Jaakkola}{2018}]{alvarez2018towards}
David Alvarez~Melis and Tommi Jaakkola.
\newblock Towards robust interpretability with self-explaining neural networks.
\newblock {\em NeurIPS}, 2018.

\bibitem[\protect\citeauthoryear{Ancona \bgroup \em et al.\egroup
  }{2017}]{ancona2017towards}
Marco Ancona, Enea Ceolini, Cengiz {\"O}ztireli, and Markus Gross.
\newblock Towards better understanding of gradient-based attribution methods
  for deep neural networks.
\newblock {\em arXiv preprint arXiv:1711.06104}, 2017.

\bibitem[\protect\citeauthoryear{Antognini and
  Faltings}{2021}]{antognini2021rationalization}
Diego Antognini and Boi Faltings.
\newblock Rationalization through concepts.
\newblock {\em arXiv:2105.04837}, 2021.

\bibitem[\protect\citeauthoryear{Arras \bgroup \em et al.\egroup
  }{2022}]{arras2022clevr}
Leila Arras, Ahmed Osman, and Wojciech Samek.
\newblock Clevr-xai: a benchmark dataset for the ground truth evaluation of
  neural network explanations.
\newblock {\em Information Fusion}, 81:14--40, 2022.

\bibitem[\protect\citeauthoryear{Bach \bgroup \em et al.\egroup
  }{2015}]{bach2015pixel}
Sebastian Bach, Alexander Binder, Gr{\'e}goire Montavon, Frederick Klauschen,
  Klaus-Robert M{\"u}ller, and Wojciech Samek.
\newblock On pixel-wise explanations for non-linear classifier decisions by
  layer-wise relevance propagation.
\newblock {\em PloS one}, 10(7):e0130140, 2015.

\bibitem[\protect\citeauthoryear{Bahadori and
  Heckerman}{2020}]{bahadori2020debiasing}
Mohammad Bahadori and David Heckerman.
\newblock Debiasing concept-based explanations with causal analysis.
\newblock {\em arXiv:2007.11500}, 2020.

\bibitem[\protect\citeauthoryear{Bau \bgroup \em et al.\egroup
  }{2017}]{bau2017network}
David Bau, Bolei Zhou, Aditya Khosla, Aude Oliva, and Antonio Torralba.
\newblock Network dissection: Quantifying interpretability of deep visual
  representations.
\newblock In {\em IEEE CVPR}, 2017.

\bibitem[\protect\citeauthoryear{Bhatt \bgroup \em et al.\egroup
  }{2020}]{bhatt2020evaluating}
Umang Bhatt, Adrian Weller, and Jos{\'e}~MF Moura.
\newblock Evaluating and aggregating feature-based model explanations.
\newblock {\em arXiv:2005.00631}, 2020.

\bibitem[\protect\citeauthoryear{Blazek and Lin}{2021}]{blazek2021explainable}
Paul Blazek and Milo Lin.
\newblock Explainable neural networks that simulate reasoning.
\newblock {\em Nature Computational Science}, 1(9):607--618, 2021.

\bibitem[\protect\citeauthoryear{Chalasani \bgroup \em et al.\egroup
  }{2020}]{chalasani2020concise}
Prasad Chalasani, Jiefeng Chen, Xi~Wu, and Somesh Jha.
\newblock Concise explanations of neural networks using adversarial training.
\newblock In {\em ICML}, 2020.

\bibitem[\protect\citeauthoryear{Chang \bgroup \em et al.\egroup
  }{2021}]{chang2021node}
Chun-Hao Chang, Rich Caruana, and Anna Goldenberg.
\newblock Neural generalized additive model for interpretable deep learning.
\newblock {\em arXiv:2106.01613}, 2021.

\bibitem[\protect\citeauthoryear{Chattopadhay \bgroup \em et al.\egroup
  }{2018}]{chattopadhay2018grad}
Aditya Chattopadhay, Anirban Sarkar, Prantik Howlader, and Vineeth
  Balasubramanian.
\newblock Grad-cam++: Generalized gradient-based visual explanations for deep
  convolutional networks.
\newblock In {\em WACV}, 2018.

\bibitem[\protect\citeauthoryear{Chauhan \bgroup \em et al.\egroup
  }{2022}]{chauhan2022interactive}
Kushal Chauhan, Rishabh Tiwari, Jan Freyberg, and Pradeep Shenoy.
\newblock Interactive concept bottleneck models.
\newblock {\em arXiv:2212.07430}, 2022.

\bibitem[\protect\citeauthoryear{Chen \bgroup \em et al.\egroup
  }{2019}]{chen2019looks}
Chaofan Chen, Oscar Li, Daniel Tao, Alina Barnett, Cynthia Rudin, and
  Jonathan~K Su.
\newblock This looks like that: deep learning for interpretable image
  recognition.
\newblock {\em NeurIPS}, 2019.

\bibitem[\protect\citeauthoryear{Chen \bgroup \em et al.\egroup
  }{2020}]{chen2020concept}
Zhi Chen, Yijie Bei, and Cynthia Rudin.
\newblock Concept whitening for interpretable image recognition.
\newblock {\em Nature Machine Intelligence}, 2(12):772--782, 2020.

\bibitem[\protect\citeauthoryear{Dasgupta \bgroup \em et al.\egroup
  }{2022}]{dasgupta2022framework}
Sanjoy Dasgupta, Nave Frost, and Michal Moshkovitz.
\newblock Framework for evaluating faithfulness of local explanations.
\newblock {\em arXiv:2202.00734}, 2022.

\bibitem[\protect\citeauthoryear{Dhurandhar \bgroup \em et al.\egroup
  }{2022}]{dhurandhar2022right}
Amit Dhurandhar, Karthikeyan Ramamurthy, and Karthikeyan Shanmugam.
\newblock Is this the right neighborhood? accurate and query efficient model
  agnostic explanations.
\newblock In {\em NeurIPS}, 2022.

\bibitem[\protect\citeauthoryear{Dubey \bgroup \em et al.\egroup
  }{2022}]{dubey2022scalable}
Abhimanyu Dubey, Filip Radenovic, and Dhruv Mahajan.
\newblock Scalable interpretability via polynomials.
\newblock {\em NeurIPS}, 2022.

\bibitem[\protect\citeauthoryear{Erion \bgroup \em et al.\egroup
  }{2021}]{erion2021improving}
Gabriel Erion, Joseph~D Janizek, Pascal Sturmfels, Scott~M Lundberg, and Su-In
  Lee.
\newblock Improving performance of deep learning models with axiomatic
  attribution priors and expected gradients.
\newblock {\em Nature MI}, 2021.

\bibitem[\protect\citeauthoryear{Fong \bgroup \em et al.\egroup
  }{2019}]{fong2019understanding}
Ruth Fong, Mandela Patrick, and Andrea Vedaldi.
\newblock Understanding deep networks via extremal perturbations and smooth
  masks.
\newblock In {\em ICCV}, 2019.

\bibitem[\protect\citeauthoryear{Ghorbani and Zou}{2020}]{ghorbani2020neuron}
Amirata Ghorbani and James~Y Zou.
\newblock Neuron shapley: Discovering the responsible neurons.
\newblock {\em NeurIPS}, 2020.

\bibitem[\protect\citeauthoryear{Hastie}{2017}]{hastie2017generalized}
Trevor Hastie.
\newblock Generalized additive models.
\newblock In {\em Statistical models}, pages 249--307. Routledge, 2017.

\bibitem[\protect\citeauthoryear{Hendricks \bgroup \em et al.\egroup
  }{2016}]{hendricks2016generating}
Lisa Hendricks, Zeynep Akata, Marcus Rohrbach, Jeff Donahue, and Trevor
  Darrell.
\newblock Generating visual explanations.
\newblock In {\em ECCV}, 2016.

\bibitem[\protect\citeauthoryear{Hooker \bgroup \em et al.\egroup
  }{2019}]{hooker2019benchmark}
Sara Hooker, Dumitru Erhan, Pieter Kindermans, and Been Kim.
\newblock A benchmark for interpretability methods in deep neural nets.
\newblock {\em NeurIPS}, 2019.

\bibitem[\protect\citeauthoryear{Jalwana \bgroup \em et al.\egroup
  }{2021}]{jalwana2021cameras}
Mohammad Jalwana, Naveed Akhtar, Mohammed Bennamoun, and Ajmal Mian.
\newblock Cameras: Enhanced resolution and sanity preserving class activation
  mapping for image saliency.
\newblock In {\em IEEE CVPR}, 2021.

\bibitem[\protect\citeauthoryear{Khakzar \bgroup \em et al.\egroup
  }{2022}]{khakzar2022explanations}
Ashkan Khakzar, Pedram Khorsandi, Rozhin Nobahari, and Nassir Navab.
\newblock Do explanations explain? model knows best.
\newblock In {\em IEEE CVPR}, 2022.

\bibitem[\protect\citeauthoryear{Kim \bgroup \em et al.\egroup
  }{2018}]{kim2018interpretability}
Been Kim, Martin Wattenberg, Justin Gilmer, Carrie Cai, James Wexler, Fernanda
  Viegas, et~al.
\newblock Interpretability beyond feature attribution: Quantitative testing
  with concept activation vectors.
\newblock In {\em ICML}, 2018.

\bibitem[\protect\citeauthoryear{Kim \bgroup \em et al.\egroup
  }{2022}]{kim2022vit}
Sangwon Kim, Jaeyeal Nam, and Byoung~Chul Ko.
\newblock Vit-net: Interpretable vision transformers with neural tree decoder.
\newblock In {\em ICML}, 2022.

\bibitem[\protect\citeauthoryear{Kindermans \bgroup \em et al.\egroup
  }{2019}]{kindermans2019reliability}
Pieter-Jan Kindermans, Sara Hooker, Julius Adebayo, Maximilian Alber, Kristof~T
  Sch{\"u}tt, Sven D{\"a}hne, Dumitru Erhan, and Been Kim.
\newblock The (un) reliability of saliency methods.
\newblock In {\em Explainable AI: IEVDL}, pages 267--280. Springer, 2019.

\bibitem[\protect\citeauthoryear{Koh \bgroup \em et al.\egroup
  }{2020}]{koh2020concept}
Pang~Wei Koh, Thao Nguyen, Yew~Siang Tang, Stephen Mussmann, Emma Pierson, Been
  Kim, and Percy Liang.
\newblock Concept bottleneck models.
\newblock In {\em ICML}, 2020.

\bibitem[\protect\citeauthoryear{Kohlbrenner \bgroup \em et al.\egroup
  }{2020}]{kohlbrenner2020towards}
Maximilian Kohlbrenner, Shinichi Bauer, Alexander Binder, and Sebastian
  Lapuschkin.
\newblock Towards best practice in explaining neural network decisions with
  lrp.
\newblock In {\em IJCNN}, 2020.

\bibitem[\protect\citeauthoryear{Kwon and Zou}{2022}]{kwon2022weightedshap}
Yongchan Kwon and James Zou.
\newblock Weighted{SHAP}: analyzing and improving shapley based feature
  attributions.
\newblock {\em arXiv:2209.13429}, 2022.

\bibitem[\protect\citeauthoryear{Lang \bgroup \em et al.\egroup
  }{2021}]{lang2021explaining}
Oran Lang, Yossi Gandelsman, Michal Yarom, Yoav Wald, Gal Elidan, Avinatan
  Hassidim, William~T Freeman, Phillip Isola, Amir Globerson, Michal Irani,
  et~al.
\newblock Explaining in style: Training a gan to explain a classifier in
  stylespace.
\newblock In {\em ICCV}, 2021.

\bibitem[\protect\citeauthoryear{Li \bgroup \em et al.\egroup
  }{2021}]{li2021scouter}
Liangzhi Li, Bowen Wang, Manisha Verma, Yuta Nakashima, Ryo Kawasaki, and
  Hajime Nagahara.
\newblock Scouter: Slot attention-based classifier for explainable image
  recognition.
\newblock In {\em ICCV}, 2021.

\bibitem[\protect\citeauthoryear{Lundberg and Lee}{2017}]{lundberg2017unified}
Scott Lundberg and Su-Im Lee.
\newblock A unified approach to interpreting model predictions.
\newblock {\em NeurIPS}, 2017.

\bibitem[\protect\citeauthoryear{Lundstrom \bgroup \em et al.\egroup
  }{2022}]{lundstrom2022rigorous}
Daniel~D Lundstrom, Tianjian Huang, and Meisam Razaviyayn.
\newblock A rigorous study of integrated gradients method and extensions to
  internal neuron attributions.
\newblock In {\em ICML}, 2022.

\bibitem[\protect\citeauthoryear{Montavon \bgroup \em et al.\egroup
  }{2018}]{montavon2018methods}
Gr{\'e}goire Montavon, Wojciech Samek, and Klaus-Robert M{\"u}ller.
\newblock Methods for interpreting and understanding deep neural networks.
\newblock {\em Digital signal processing}, 73:1--15, 2018.

\bibitem[\protect\citeauthoryear{Mu and Andreas}{2020}]{mu2020compositional}
Jesse Mu and Jacob Andreas.
\newblock Compositional explanations of neurons.
\newblock {\em NeurIPS}, 2020.

\bibitem[\protect\citeauthoryear{Nguyen and
  Mart{\'\i}nez}{2020}]{nguyen2020quantitative}
An-phi Nguyen and Mar{\'\i}a~Rodr{\'\i}guez Mart{\'\i}nez.
\newblock On quantitative aspects of model interpretability.
\newblock {\em arXiv:2007.07584}, 2020.

\bibitem[\protect\citeauthoryear{Omeiza \bgroup \em et al.\egroup
  }{2019}]{omeiza2019smooth}
Daniel Omeiza, Skyler Speakman, Celia Cintas, and Komminist Weldermariam.
\newblock An enhanced inference level visualization technique for deep
  convolutional neural network models.
\newblock {\em arXiv:1908.01224}, 2019.

\bibitem[\protect\citeauthoryear{Oquab \bgroup \em et al.\egroup
  }{2015}]{oquab2015object}
Maxime Oquab, L{\'e}on Bottou, Ivan Laptev, and Josef Sivic.
\newblock Is object localization for free?-weakly-supervised learning with
  convolutional neural networks.
\newblock In {\em IEEE CVPR}, 2015.

\bibitem[\protect\citeauthoryear{Park \bgroup \em et al.\egroup
  }{2018}]{park2018multimodal}
Dong Park, Lisa~Anne Hendricks, Zeynep Akata, Anna Rohrbach, Bernt Schiele,
  Trevor Darrell, and Marcus Rohrbach.
\newblock Multimodal explanations: Justifying decisions and pointing to the
  evidence.
\newblock In {\em CVPR}, 2018.

\bibitem[\protect\citeauthoryear{Petsiuk \bgroup \em et al.\egroup
  }{2018}]{petsiuk2018rise}
Vitali Petsiuk, Abir Das, and Kate Saenko.
\newblock Rise: Randomized input sampling for explanation of black-box models.
\newblock {\em arXiv:1806.07421}, 2018.

\bibitem[\protect\citeauthoryear{Rao \bgroup \em et al.\egroup
  }{2022}]{rao2022towards}
Sukrut Rao, Moritz B{\"o}hle, and Bernt Schiele.
\newblock Towards better understanding attribution methods.
\newblock In {\em IEEE CVPR}, 2022.

\bibitem[\protect\citeauthoryear{Ribeiro \bgroup \em et al.\egroup
  }{2016}]{ribeiro2016should}
Marco~Tulio Ribeiro, Sameer Singh, and Carlos Guestrin.
\newblock " why should i trust you?" explaining the predictions of any
  classifier.
\newblock In {\em ACM SIGKDD}, 2016.

\bibitem[\protect\citeauthoryear{Rong \bgroup \em et al.\egroup
  }{2022}]{rong2022consistent}
Yao Rong, Tobias Leemann, Vadim Borisov, Gjergji Kasneci, and Enkelejda
  Kasneci.
\newblock A consistent and efficient evaluation strategy for attribution
  methods.
\newblock In {\em ICML}, 2022.

\bibitem[\protect\citeauthoryear{Rudin}{2019}]{rudin2019stop}
Cynthia Rudin.
\newblock Stop explaining black box machine learning models for high stakes
  decisions and use interpretable models instead.
\newblock {\em NMI}, 1(5):206--215, 2019.

\bibitem[\protect\citeauthoryear{Samek \bgroup \em et al.\egroup
  }{2016}]{samek2016evaluating}
Wojciech Samek, Alexander Binder, Gr{\'e}goire Montavon, and Klaus-Robert
  M{\"u}ller.
\newblock Evaluating the visualization of what a deep neural network has
  learned.
\newblock {\em IEEE TNNLS}, 28(11):2660--2673, 2016.

\bibitem[\protect\citeauthoryear{Sarkar \bgroup \em et al.\egroup
  }{2022}]{sarkar2022framework}
Anirban Sarkar, Deepak Vijaykeerthy, Anindya Sarkar, and Vineeth~N
  Balasubramanian.
\newblock A framework for learning ante-hoc explainable models via concepts.
\newblock In {\em IEEE CVPR}, 2022.

\bibitem[\protect\citeauthoryear{Selvaraju \bgroup \em et al.\egroup
  }{2017}]{selvaraju2017grad}
Ramprasaath Selvaraju, Michael Cogswell, Abhishek Das, Devi Parikh, and Dhruv
  Batra.
\newblock Grad-cam: Visual explanations from deep networks via gradient-based
  localization.
\newblock In {\em ICCV}, 2017.

\bibitem[\protect\citeauthoryear{Shapley}{1997}]{shapley1997value}
Lloyd~S Shapley.
\newblock A value for n-person games.
\newblock {\em Classics in game theory}, 69, 1997.

\bibitem[\protect\citeauthoryear{Shrikumar \bgroup \em et al.\egroup
  }{2017}]{shrikumar2017learning}
Avanti Shrikumar, Peyton Greenside, and Anshul Kundaje.
\newblock Learning important features through propagating activations.
\newblock In {\em ICML}, 2017.

\bibitem[\protect\citeauthoryear{Simonyan \bgroup \em et al.\egroup
  }{2014}]{simonyan2013deep}
Karen Simonyan, Andrea Vedaldi, and Andrew Zisserman.
\newblock Deep inside convolutional networks: Visualising image classification
  models and saliency maps.
\newblock {\em ICLR}, 2014.

\bibitem[\protect\citeauthoryear{Sixt \bgroup \em et al.\egroup
  }{2020}]{sixt2020explanations}
Leon Sixt, Maximilian Granz, and Tim Landgraf.
\newblock When explanations lie: Why many modified bp attributions fail.
\newblock In {\em ICML}, 2020.

\bibitem[\protect\citeauthoryear{Smilkov \bgroup \em et al.\egroup
  }{2017}]{smilkov2017smoothgrad}
Daniel Smilkov, Nikhil Thorat, Been Kim, and Martin Wattenberg.
\newblock Smoothgrad: removing noise by adding noise.
\newblock {\em arXiv:1706.03825}, 2017.

\bibitem[\protect\citeauthoryear{Springenberg \bgroup \em et al.\egroup
  }{2015}]{springenberg2014striving}
Jost Springenberg, Alexey Dosovitskiy, Thomas Brox, and Martin Riedmiller.
\newblock Striving for simplicity: The all convolutional net.
\newblock {\em ICLR}, 2015.

\bibitem[\protect\citeauthoryear{Srinivas and Fleuret}{2019}]{srinivas2019full}
Suraj Srinivas and Fran{\c{c}}ois Fleuret.
\newblock Full-gradient representation for neural network visualization.
\newblock {\em NeurIPS}, 2019.

\bibitem[\protect\citeauthoryear{Stalder \bgroup \em et al.\egroup
  }{2022}]{stalder2022you}
Steven Stalder, Nathana{\"e}l Perraudin, Radhakrishna Achanta, Fernando
  Perez-Cruz, and Michele Volpi.
\newblock What you see is what you classify: Black box attributions.
\newblock {\em NeurIPS}, 2022.

\bibitem[\protect\citeauthoryear{Sturmfels \bgroup \em et al.\egroup
  }{2020}]{sturmfels2020visualizing}
Pascal Sturmfels, Scott Lundberg, and Su-In Lee.
\newblock Visualizing the impact of feature attribution baselines.
\newblock {\em Distill}, 2020.

\bibitem[\protect\citeauthoryear{Sundararajan \bgroup \em et al.\egroup
  }{2017}]{sundararajan2017axiomatic}
Mukund Sundararajan, Ankur Taly, and Qiqi Yan.
\newblock Axiomatic attribution for deep networks.
\newblock In {\em ICML}, 2017.

\bibitem[\protect\citeauthoryear{Tomsett \bgroup \em et al.\egroup
  }{2020}]{tomsett2020sanity}
Richard Tomsett, Dan Harborne, Supriyo Chakraborty, Prudhvi Gurram, and Alun
  Preece.
\newblock Sanity checks for saliency metrics.
\newblock In {\em AAAI}, 2020.

\bibitem[\protect\citeauthoryear{Wang \bgroup \em et al.\egroup
  }{2020}]{wang2020score}
Haofan Wang, Zifan Wang, Mengnan Du, Fan Yang, Zijian Zhang, Sirui Ding, Piotr
  Mardziel, and Xia Hu.
\newblock Score-weighted visual explanations for convolutional neural networks.
\newblock In {\em IEEE CVPRw}, 2020.

\bibitem[\protect\citeauthoryear{Wang \bgroup \em et al.\egroup
  }{2021}]{wang2021interpretable}
Jiaqi Wang, Huafeng Liu, Xinyue Wang, and Liping Jing.
\newblock Interpretable image recognition by constructing transparent embedding
  space.
\newblock In {\em ICCV}, 2021.

\bibitem[\protect\citeauthoryear{Wang \bgroup \em et al.\egroup
  }{2022}]{wang2022hint}
Andong Wang, Wei-Ning Lee, and Xiaojuan Qi.
\newblock Hint: Hierarchical neuron concept explainer.
\newblock In {\em IEEE CVPR}, 2022.

\bibitem[\protect\citeauthoryear{Yang \bgroup \em et al.\egroup
  }{2023}]{yang2023local}
Peiyu Yang, Naveed Akhtar, Zeyi Wen, and Ajmal Mian.
\newblock Local path integration for attribution.
\newblock In {\em AAAI Conference on Artificial Intelligence}. AAAI, 2023.

\bibitem[\protect\citeauthoryear{Yeh \bgroup \em et al.\egroup
  }{2019}]{yeh2019fidelity}
Chih-Kuan Yeh, Cheng-Yu Hsieh, Arun Suggala, David Inouye, and Pradeep
  Ravikumar.
\newblock On the (in)fidelity and sensitivity of explanations.
\newblock {\em NeurIPS}, 2019.

\bibitem[\protect\citeauthoryear{Zhang \bgroup \em et al.\egroup
  }{2018}]{zhang2018top}
Jianming Zhang, Sarah~Adel Bargal, Zhe Lin, Jonathan Brandt, Xiaohui Shen, and
  Stan Sclaroff.
\newblock Top-down neural attention by excitation backprop.
\newblock {\em IJCV}, 126(10):1084--1102, 2018.

\bibitem[\protect\citeauthoryear{Zhou \bgroup \em et al.\egroup
  }{2016}]{zhou2016learning}
Bolei Zhou, Agata Lapedriza, Aude Oliva, and Antonio Torralba.
\newblock Learning deep features for discriminative localization.
\newblock In {\em IEEE CVPR}, 2016.

\end{thebibliography}

\end{document}